\documentclass[a4paper, 10pt, conference]{ieeeconf}      

\IEEEoverridecommandlockouts                              

\usepackage[usestackEOL]{stackengine}

\usepackage{gensymb}
\usepackage{graphicx}
\usepackage{subcaption}
\usepackage{xcolor}
\usepackage{amsmath}
\usepackage{amssymb}
\usepackage{float}
\usepackage{cite}
\usepackage{multirow}
\usepackage{authblk}
\usepackage{tabularx}
\usepackage{placeins}
\usepackage[margin=0.76in]{geometry}
\usepackage{caption}
\usepackage[T1]{fontenc}
\usepackage[utf8]{inputenc}
\usepackage{authblk}
\title{\LARGE \bf
Learning How to Trade-Off Safety with Agility Using Deep Covariance Estimation for Perception Driven UAV Motion Planning
}

\author{%
Onur Akgun$^{1^*}$, Kamil Canberk Atik$^{2^*}$, Mustafa Erdem$^{3}$, Mehmetcan Kaymaz$^{2}$, Bugrahan Yamak$^{2}$, \\ and N. Kemal Ure$^{2}$
\thanks{*These authors contributed equally to this work}
\thanks{$^{1}$Faculty of Engineering, Mechatronics Engineering, Turkish-German University, Istanbul, Turkey
        {\tt\small akgun@tau.edu.tr}}
\thanks{$^{2}$Faculty of Aeronautics and Astronautics and ITU Artificial Intelligience and Data Science Application and Research Center, Istanbul Technical University, Istanbul, Turkey,
        {\tt\small atik20, kaymazm16, yamak18,ure@itu.edu.tr, }}
\thanks{$^{3}$Graduate School of Science, Engineering and Technology, Mechatronics Engineering, Istanbul Technical University, Istanbul, Turkey,
        {\tt\small erdemm@itu.edu.tr}}
}

\makeatletter
\def\endthebibliography{%
  \def\@noitemerr{\@latex@warning{Empty `thebibliography' environment}}%
  \endlist
}
\makeatother

\begin{document}

\maketitle
\thispagestyle{empty}
\pagestyle{empty}


\begin{abstract}
We investigate how to utilize predictive models for selecting appropriate motion planning strategies based on perception uncertainty estimation for agile unmanned aerial vehicle (UAV) navigation tasks. Although there are variety of motion planning and perception algorithms for such tasks, the impact of perception uncertainty is not explicitly handled in many of the current motion algorithms, which leads to performance loss in real-life scenarios where the measurement are often noisy due to external disturbances. We develop a novel framework for embedding perception uncertainty to high level motion planning management, in order to select the best available motion planning approach for the currently estimated perception uncertainty. We estimate the uncertainty in visual inputs using a deep neural network (CovNet) that explicitly predicts the covariance of the current measurements. Next, we train a high level machine learning model for predicting the lowest cost motion planning algorithm given the current estimate of covariance as well as the UAV states. We demonstrate on both real-life data and drone racing simulations that our approach, named uncertainty driven motion planning switcher (UDS) yields the safest and fastest trajectories among compared alternatives. Furthermore, we show that the developed approach learns how to trade-off safety with agility by switching to motion planners that leads to more agile trajectories when the estimated covariance is high and vice versa.
\end{abstract}

\section{Introduction}
Agile unamanned aerial vehicle (UAV) navigation tasks, such as drone racing with first-person view (FPV) cameras, has been gaining popularity recent years. In particular, autonomous drone racing has become an attractive topic for robotic researchers\cite{foehn2020alphapilot}, due to its unique setting that combines agile motion planning, advanced perception and nonlinear control. All the aforementioned systems are required to run on on-board hardware with very limited processing power and memory capacity\cite{moon2019challenges}, which further increases the level of challenge. Although there has been significant improvements in agile motion planning and autonomous drone racing research, current systems' performance are still not a match to their human-controlled counterparts\cite{kaufmann2018deep}. 


\begin{figure}
    \centering
    \includegraphics[width=0.5\textwidth]{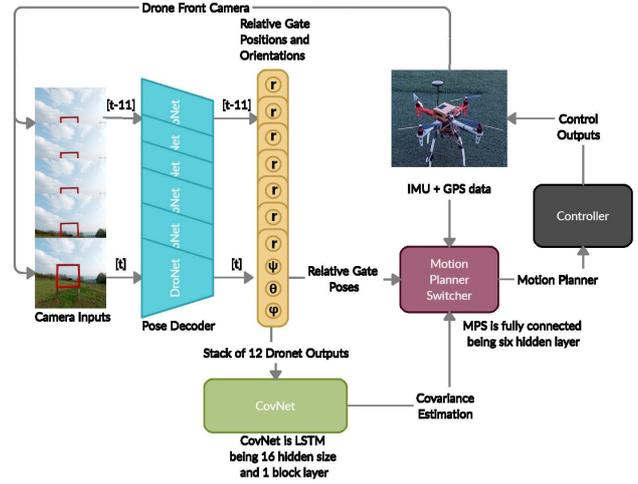}
    \caption{Overview of the Uncertainty Driven Motion Planner Switcher (UDS). We use a sequence of visual inputs and novel deep neural network architecture (CovNet) to estimate covariance of the relative gate pose estimations. A pre-trained motion planning switcher then predicts the lowest cost motion planning strategy for the current covariance and state estimates.}
    \label{fig:wholesystem}
\end{figure}

Developing a robust, accurate and fast perception system is one of the central challenges in agile UAV motion planning. Recent advances in deep learning based computer vision enabled unprecedented improvements in this area, hence deep neural networks are now a common part of drone perception systems\cite{tai2016deep}. However, even the most advanced deep learning systems fail to operate robustly in noisy environments\cite{min2018survey}. In particular, motion blur due to agile maneuvering of the UAV or external light sources that negatively effect camera exposure can lead to high error rates. Since the performance of motion planning and perception is tightly coupled in this setting\cite{pendleton2017perception}, these high error rates can lead to significant performance loss in motion planning module. Over-relying on perception performance often results in agile maneuvers that violate safety conditions and results in crash/loss of the UAV. On the other hand, having low confidence in perception output leads to maneuvers with limited speed and acceleration, which in turn negatively affects mission metrics such as lap time. Although there has been previous attempts at integrating perception and motion planning modules\cite{lu2018survey}, developing an integrated system that can trade-off safety with agility is still an open problem. The main objective of this paper is to develop a such system through explicit estimation of perception uncertainty and using these estimations to adapt the motion planning strategy on the fly.

\subsection{Contributions}
Our overall approach, named Uncertainy Driven Motion Planner Switcher (UDS) is summarized in Fig. \ref{fig:wholesystem}.~The contributions of our work can be summarized as follows:
\begin{itemize}
    \item \textbf{Explicit estimation of perception covariance for relative gate pose predictions}: We extend the existing DroNet\cite{dronet} architecture with additional recurrent layers to estimate the covariance of relative gate pose  predictions based in visual inputs. These additional layers and the customized training procedure allows us to quantify the uncertainty in the neural network predictions. We verify our approach on real life data, and show that the system can reliably output low covariance estimates for clean visual inputs and high covariance estimates under external disturbances such as excess blur and composure. Details of the covariance estimation process is provided in Section \ref{s:covnet}.
    \item \textbf{A switched motion planning framework for adapting to measurement uncertainty}: We develop a motion planning framework that can switch between various different motion planning strategies based on the current UAV states and covariance estimates. In order to find a switching strategy that does not rely on heuristics, we develop a data-driven approach, where we train a fully connected neural network to predict which motion planning approach would result in best performance under given noise/covariance estimates. We verify that the trained network learns to select more agile motion planning approaches in less noisy scenarios and smoother/safer motion planning approaches in high covariance/noisy settings. Details of the motion planning approach is provided in Section \ref{s:uds}.
\end{itemize}
Finally, we demonstrate the capabilities of the integrated perception-motion planning system on a simulated drone racing scenario (see Section). We show that UDS achieves the best performance compared to alternative approaches. We highlight that this result is due to systems' ability to achieve trade-off between safety and agility under exposition to different noise levels. 


\subsection{Related Work}
\subsubsection{Perception}
Traditional visual inertial odometry (VIO) and simultaneous localization and mapping (SLAM) methods used in UAV perception systems are include~\cite{msckf,iterated,linear} for VIO and \cite{orbslam,maplabslam} for SLAM. Although these methods are theoretically sound, they suffer from high computational complexity, lack of operational speed, lack of robustness to camera noise, and need of extensive calibration for camera and IMU. Modern approaches to VIO mostly relies on deep neural networks~\cite{vinet}, which offer improved robustness and accuracy. There are also alternative methods which only rely on visual inputs for navigation~\cite{undeepvo,gen}, which are less accurate/robust, but are faster and do not require any other sensor than camera. Methods based on event-based cameras~\cite{event} are also gaining popularity, however they are still in their early adoption phase. 
In autonomous drone racing, gate pose detection is another important perception problem, since precise determination of relative gate distance and orientation is crucial to completing the race. For this problem, there are both methods which require a prior map of the environment\cite{8793631}, and methods that do not require a map, such as snake gate detection\cite{delft}, end-to-end learnable visuomotor policies\cite{microsoft} and gap detection \cite{gapflyt}.  
\subsubsection{Motion Planning}
Motion planning is another critical system for agile UAV navigation tasks. According to \cite{gonzalez2015review},  motion planning algorithms are classified in four groups: graph search, interpolating, numerical optimization and sampling. 
In \cite{8206119}, authors have proposed  a graph search based planning method to compute feasible smooth,  minimum-time trajectories for a quadrotor. Kayacan E. et. al. used sampling based motion planning algorithm in order to solve formation landing problem of quadcopter \cite{7838567}. \cite{gebhardt2016airways} is an example of optimization based trajectory planning for quadrotors. In this study, user provided rough keyframes are used to generate 3D feasible trajectories via optimization based methods. There are also perception aware motion planning studies. In \cite{costante2016perception}, authors have developed a solution to localization uncertainty by combining perception and motion planning.
\subsubsection{Drone Racing}
Autonomous drone racing also has been heavily dominated by neural network based methods. In a recent end-to-end method, there are two neural structures are applied, first of which is to imitate the behaviors of controller and motion planner, and the second one is a reinforcement learning block for fine tuning of previously trained model \cite{narrowgap}. In another  vision based approach, two neural networks are employed so that both the distance between the gate and drone and its covariance values could be calculated~\cite{8793631} based on a previously supplied map. Note that our method does not require such a map and estimates convariance by only using sequence of visual inputs.

Another study maps neural network outputs to motion planner, which generates minimum-jerk trajectory to reach the desired goal\cite{kaufmann2018deep}. 

\section{Deep Covariance Estimation}\label{s:covnet}

Gate pose prediction is one of the main tasks of drone racing. In this subsection, we provide the design of a deep neural network based system, where in addition to standard pose estimates (relative distance and orientation between UAV and the gate) also outputs the 
covariance of each of these estimates. The covariance estimates are later used to quantify the uncertainty of the perception system and drive the motion planning switching strategy.

\subsection{Frame Convention}
Inertial frame is represented by $I$, body frame is represented by $B$ and the gate frame is represented by $G$. $P_{I_B}$ is position and orientation of drone with respected to inertial frame. $P_{I_G}$ is position and orientation of the gate with respect to inertial frame. For calculating the gate location and orientation in inertial frame, our proposed network takes measurements in $P_{B_G}$, which is position and orientation of gate in drone body frame.     

$P_{I_B}$ and $P_{I_G}$ are parameterized by a Cartesian coordinate system, with variables $(x,y,z)$. For the training, $P_{I_B}$ data is provided as ground truth in simulation and estimated with flight control unit. However, $P_{I_G}$ data is calculated by on-board system based on perception system data and flight control data. $P_{GB}^b$ is parameterized in polar coordinates, in body frame $(r,\theta,\psi)$. 
$P_{GB}^b$, predicted by the network, is converted to polar coordinate to Cartesian coordinate with Eq. 1., which is then transformed from $P_{GB}^b$ to $P_{GB}^i$. Finally, the sum of $P_{GB}^i$ and $-P_{I_B}$ is $P_{I_G}$, which is the final output.
\begin{equation}
    \begin{aligned}
        x &= r \sin\theta \sin\psi\\
        y &= r \sin\theta \cos\psi\\
        z &= r \cos\psi
    \end{aligned}
\end{equation}

\subsection{Perception System}

	\subsubsection{System Architecture}
	
	Perception system consists of the left side of Fig. \ref{fig:wholesystem}, whose inputs are $200\times200$ RGB images, and the output is a $4$ dimensional vector  $\hat{y}_{B_G} = [\hat{r},\hat{\psi},\hat{\theta},\hat{\phi}] \in \mathbb{R}^4$, $ \mathbb{R}^4 \in SO(3)$, which is the predicted gate pose. $\hat{r}$ is the distance between drone body frame origin and gate frame origin. $\hat{\psi}$ and $\hat{\theta}$ are relative orientation between drone and the gate in spherical coordinates. After calculating difference of gate and body frame on Cartesian coordinate system, this result is converted to spherical coordinates via Eq. \ref{e:convert}. $\hat{\phi}$ is the yaw angle difference between drone body frame and the gate frame. 
	\begin{equation}
    	\begin{aligned}
    	    r &= \sqrt{x^2 + y^2 + z^2}\\
    	    \psi &= \arctan{\frac{y}{x}}\\
    	    \theta &= \arctan{\frac{\sqrt{x^2 + y^2}}{r}}
    	\end{aligned}\label{e:convert}
	\end{equation}
	
	Our proposed perception system consists of two different architectures. We use the existing DroNet\cite{dronet}, which is a convolutional neural network (CNN) based network with $8$ residual layers, to predict gate poses $\hat{y}_{B_G}$ from single input images. As shown in Fig. \ref{fig:wholesystem}, we feed a sequential batch of $12$ images to DroNet, and collect the predicted gate poses for each frame.
     \begin{figure}
        \centering
            \includegraphics[width=0.5\textwidth]{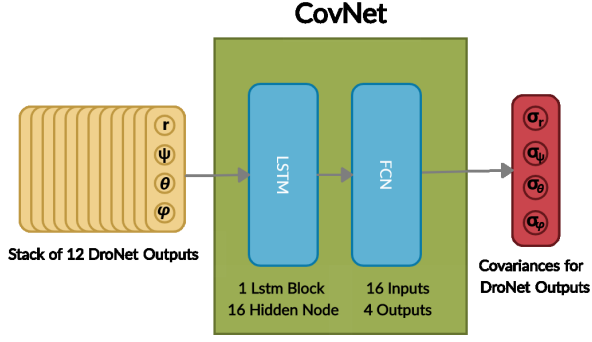}
        \caption{CovNet Network Architecture: LSTM and FCN layers are combined and trained to estimate covariance of predicted values of the gates pose variables predicted by the perception system.
        }
        \label{fig:CovNet}
    \end{figure}
	The second architecture, and the novel part of our  overall system is CovNet, shown in Fig. \ref{fig:CovNet}. CovNet's inputs are the $12$ gate pose predictions provided by DroNet. CovNet uses long-short-term memory (LSTM) and fully connected (FCN) layers to process these inputs and outputs covariance estimates for each predicted gate pose variable. This architecture is motivated by the fact that, whenever camera is exposed to high noise levels, at least several of the $12$ DroNet outputs are expected to be inconsistent with each other. Hence the anomaly in this sequential DroNet predictions can be detected by LSTM layers, which is later can be used to estimate covariance of these predictions. 

    \subsubsection{Training Data Generation and Procedure}

    We train the aforementioned DroNet and CovNet networks using the AirSim simulator \cite{shah2018airsim}. We generate around $175,000$ sequence of images using various gate and drone configurations. First, we train DroNet in isolation. The loss function used in DroNet training is given in Eq. \ref{e:dronet}, which is simply the mean squared error between true and predicted gate pose values. For training the network, we use stochastic gradient descent with learning rate set to $10^{-4}$, batch size set to $16$, weight decay set to $10^{-2}$ and number of training epochs set to $100$.
    \begin{equation}\label{e:dronet}
    L\left(\theta_{DroNet}\right) =  \frac{1}{N} \sum_{n=0}^{N} ||y_n - \hat{y}_{BG_n}||^2 
    \end{equation}
    For training CovNet, we construct the covariance labels by computing the squared error between DroNet predictions and ground truth values provided by Airsim, as outlined in Eq. \ref{e:cov}. The loss for Covnet is the mean squared error between CovNet predictions $\tilde{R_n}$ and the covariance labels $R_n$.

    \begin{equation}\label{e:cov}
        \begin{aligned}
        \tilde{e_n} &= y_n - \tilde{y}_{B_{G_n}}  \\
            R_n &= diag([\tilde{e_n}.\tilde{e_n}^T]) \\
             &= [\sigma^2_{r}, \sigma^2_{\phi}, \sigma^2_{\theta}, \sigma^2_{\psi}]
        \end{aligned}
    \end{equation}
     \begin{equation}\label{e:covnet}
    L\left(\theta_{CovNet}\right) =  \frac{1}{N} \sum_{n=0}^{N} ||R_n - \hat{R}_{n}||^2 
    \end{equation}

    Training parameters for the CovNet are, $0.0005$, $2$, $0.01$ for learning rate, batch size and weight decay correspondingly. The training process converged in around $45$ epochs with a loss of $~0.21$.

\section{Uncertainty Driven Motion Planner Switching}\label{s:uds}
In this Section, we provide details on our switch motion planning approach, where we train classifier to predict which motion planning strategy to use given the current estimates provided by DroNet and CovNet networks developed in the previous Section.

\subsection{Motion Planning Strategies}
We use $4$ different motion planning strategies as a part of our motion planning library. Although one can pick any number of different strategies, these $4$ are the most common ones in previous work and each presents a different balance between agility and smoothness of the maneuver. These $4$ strategies correspond to four different polynomial interpolation trajectories were determined named: 1)minimum velocity, 2)minimum acceleration, 3)minimum jerk and 4)minimum jerk with full stop. The only difference between minimum jerk and minimum jerk with full stop is goal configuration in terms of velocity and acceleration. Therefore, only one of them was depicted here. The polynomial expressions that define planners were shared in Eqs. \ref{eq:f1}, \ref{eq:f2}, \ref{eq:f3} in matrix forms. 
\begin{equation}
\begin{bmatrix}
p_{i} \\
p_{f} \\
\end{bmatrix} = \begin{bmatrix}
1 & t_{i} \\
1 & t_{f} \\
\end{bmatrix}\begin{bmatrix}
c_{0} \\
c_{1} \\
\end{bmatrix}
\label{eq:f1}
\end{equation}
\begin{equation}
\begin{bmatrix}
p_{i} \\
v_{i} \\
p_{f} \\
v_{f}
\end{bmatrix} = \begin{bmatrix}
1 & t_{i} & t_{i}^{2} & t_{i}^{3} \\
0 & 1 & 2 t_{i} & 3 t_{i}^{2} \\
1 & t_{f} & t_{f}^{2} & t_{f}^{3} \\
0 & 1 & 2 t_{f} & 3 t_{f}^{2}
\end{bmatrix}\begin{bmatrix}
c_{0} \\
c_{1} \\
c_{2} \\
c_{3}
\end{bmatrix}
\label{eq:f2}
\end{equation}
\begin{equation}
\begin{bmatrix}
p_{i} \\
v_{i} \\
a_{i} \\
p_{f} \\
v_{f} \\
a_{f}
\end{bmatrix} = \begin{bmatrix}
1 & t_{i} & t_{i}^{2} & t_{i}^{3} & t_{i}^{4} & t_{i}^{5} \\
0 & 1 & 2 t_{i} & 3 t_{i}^{2} & 4 t_{i}^{3} & 5 t_{i}^{4} \\
0 & 0 & 2 & 6 t_{i} & 12 t_{i}^{2} & 20 t_{i}^{3} \\
1 & t_{f} & t_{f}^{2} & t_{f}^{3} & t_{f}^{4} & t_{f}^{5} \\
0 & 1 & 2 t_{f} & 3 t_{f}^{2} & 4 t_{f}^{3} & 5 t_{f}^{4} \\
0 & 0 & 2 & 6 t_{f} & 12 t_{f}^{2} & 20 t_{f}^{3}
\end{bmatrix}\begin{bmatrix}
c_{0} \\
c_{1} \\
c_{2} \\
c_{3} \\
c_{4} \\
c_{5}
\end{bmatrix}
\label{eq:f3}
\end{equation}

\begin{figure*}
\centering
\subfloat[Position]{\includegraphics[width=.33\linewidth]{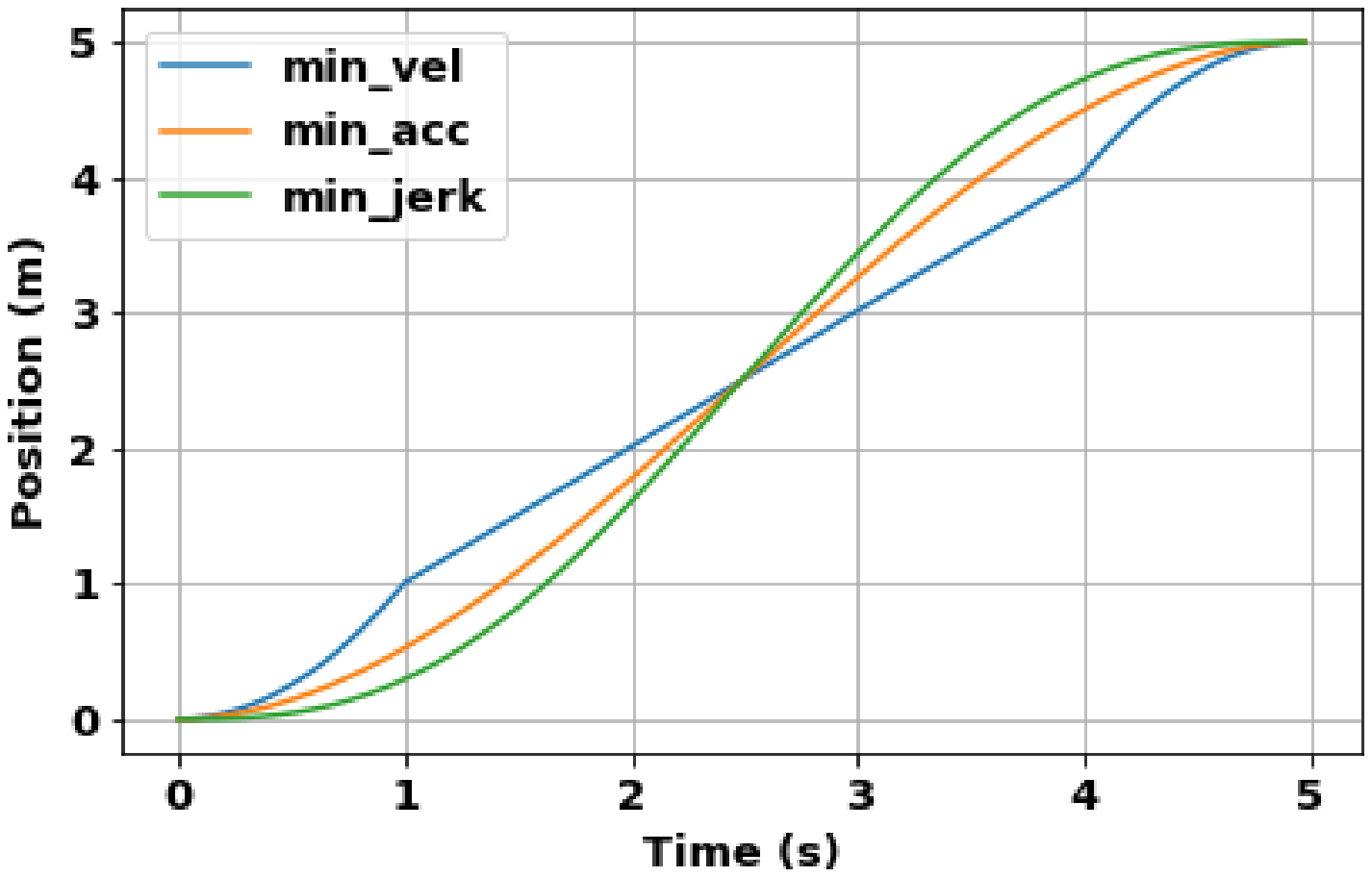}}
\hfil
\subfloat[Velocity]{\includegraphics[width=.33\linewidth]{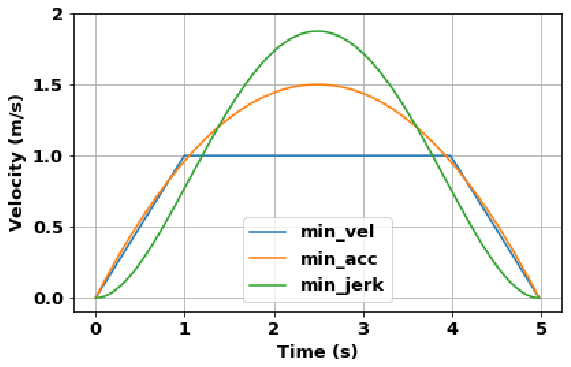}}
\hfil
\subfloat[Acceleration]{\includegraphics[width=.33\linewidth]{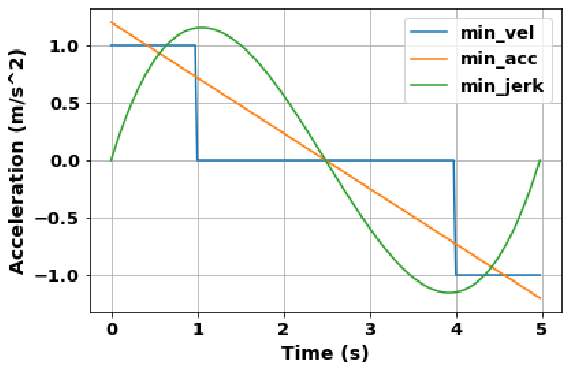}}
\caption{Trajectory characteristics of the different motion planning strategies used.} \label{fig:chars}
\end{figure*}

Here,
\begin{itemize}
    \item $t_i$, $t_f$ initial and final time
    \item $p_i$, $p_f$ initial and final position
    \item $v_i$, $v_f$ initial and final velocity
    \item $a_i$, $a_f$ initial and final acceleration
    \item $c_0, c_1, ... c_5$  are the polynomial coefficients
\end{itemize}
Fig. \ref{fig:chars} illustrates trajectory characteristics of these different strategies in a single axis. After determination of initial and final configuration parameters(time,position etc.) for the corresponding planner, the polynomial coefficients that are used to define intermediate points of flight can be calculated by simply solving corresponding linear equations.

\subsection{Classifier design for predicting the lowest cost motion planning strategy}
\begin{figure}
    \centering
\includegraphics[scale=0.3]{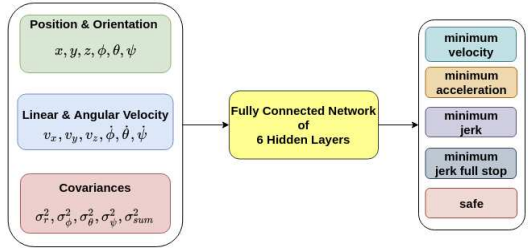}
\caption{Classifier architecture for UDS. This neural network takes drone states and covariance estimates as inputs and predicts the lowest cost motion planning strategy.}
    \label{fig:nn_structure}
\end{figure}
Selecting the appropriate/lowest cost motion planning given the current flight conditions is an important high level planning task. For instance, when the perception uncertainty is high, a less agile motion planning strategy such as minimum velocity could be used, in order to avoid possible collisions. In cases where the uncertainty is extremely high, it might be safer to switch to a safe mode where the drone stays in hover position. On the other hand, when the estimated uncertainty is low, more agile maneuvers can be executed with high confidence. A such high-level motion planning strategy could be designed with heuristics, however, in order to avoid bias and improve scalability, we use a data-driven approach, where we train a classifier to predict the lowest cost motion planning strategy given the current uncertainty estimates.

We choose a deep neural network with fully connected layers for out classifier architecture, as seen in Fig.\ref{fig:nn_structure}. For training the classifier, we select $17$ features, which are 
\begin{itemize}
    \item Difference between positions of nearest gate and drone ($x, y, z$, in meters)
    \item Difference between rotations of nearest gate and drone ($\phi, \theta, \psi$, in radians)
    \item Linear and angular velocities of drone at corresponding moment ($v_x, v_y, v_z$ in m/s and $\dot{\phi}, \dot{\theta}, \dot{\psi}$ in rad/s)
    \item Covariance measurements and its summation ($\sigma^2_{r}, \sigma^2_{\phi}, \sigma^2_{\theta}, \sigma^2_{\psi}, \sigma^2_{sum}$)
\end{itemize}
The classifier outputs class probabilities for $5$ classes, which are \textit{minimum velocity, minimum acceleration, minimum jerk, minimum jerk full stop} and \textit{safe mode}. Four of which are classical motion planning algorithms, but Safe mode, on the other hand, indicates that in case of intense noisy measurements in Figure \ref{fig:frames}, drone should stay in hover position at that moment. With the help of this option, drone can avoid to move in severe circumstances. As it will be demonstrated in Section~\ref{s:res}, this option provides a safe flight despite the agile movements during the mission. 

Cost function for the currently selected motion planning strategy is defined as follows:
\begin{equation*}
    \resizebox{1.0\hsize}{!}{$\mathcal{L} = \sqrt{(x_d - x)^2 + (y_d - y)^2 + (z_d - z)^2 + (|\psi_d - \psi| - \frac{\pi}{2})^2} . T_{arrival}$},
\end{equation*}
where $x_d, y_d, z_d$ and $\phi_d$ represent the desired location and yaw angle respectively. Additionally, $T_{arrival}$ shows the time taken to reach the desired position and direction. This function measures, how far the drone is to the desired position and orientation.

In order to collect data for training the classifier, drone was initialized with random initial states and asked to detect gates using the perception system detailed in Section \ref{s:covnet}, and navigate through them with all planners one at a time. In order to simulate different noise conditions, we expose drone camera to varying level of brightness conditions. The planner that has minimum cost was used as the correct label for corresponding features. With this approach, a dataset with approximately $120,000$ input-output pairs has been created.  
Table \ref{table:dataset} summarizes data usage and classification accuracy results. As seen in Table \ref{table:dataset} a test accuracy of $~88\%$ is achieved, which shows that it is possible to predict the lowest cost motion planning strategy with high accuracy. Once the classifier is trained, we proceed to the next step, which is selecting the planner. We simply choose the planning method which has the highest class probability. For the low level control system, we use a variant of backstepping design to track the outputs of the motion planning strategy. Further implementation details regarding the controller desing can be found in~\cite{8266387}.

\begin{table}
\begin{center}
\label{tbl:data_acc_loss}
\begin{tabular}{|l|l|l|l|}
\cline{1-4}
 Type & Data Size  & Accuracy & Loss\\ \cline{1-4}
 Train & (78549,17) & 0.903 & 0.212 \\ \cline{1-4}
 Validation & (26184,17) & 0.887 & 0.279 \\ \cline{1-4}
 Test & (15000,17) & 0.883 & 0.275 \\ \cline{1-4}
\end{tabular}
\caption{Classification statistics for the motion planning strategy prediction for train, test and validation sets.}
\label{table:dataset}
\end{center}
\end{table}

\section{Results}\label{s:res}
We provide two different test cases for validating our approach. First, we take advantage of using a simulated environment to showcase UDS in a drone racing setting with multiple gates and emulated noise. Second, we demonstrate the performance of CovNet in a real world setting, where the network estimates the covariance/uncertainty of the predictions based on visual inputs of real gate images.
\subsection{Simulation Experiments}
\begin{figure*}
\centering
\subfloat[]{\includegraphics[width=.2\textwidth]{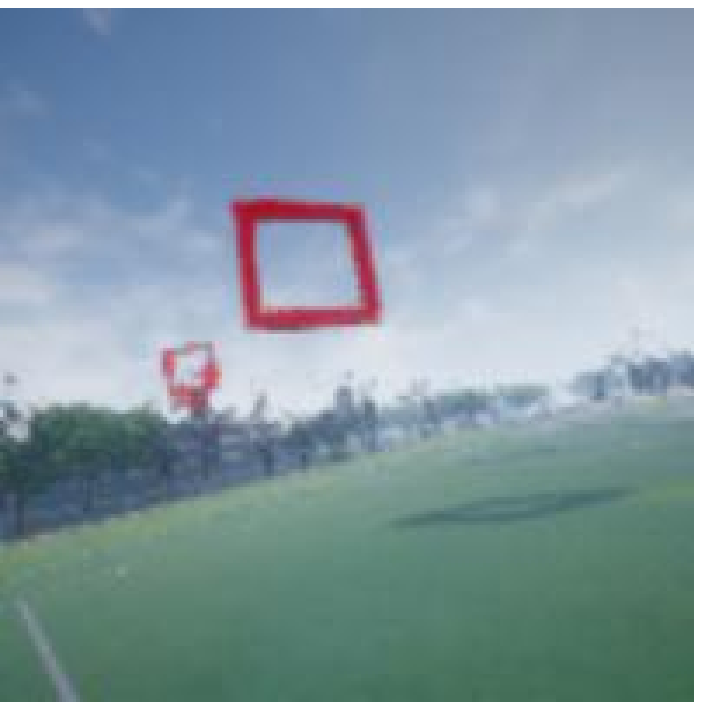} \label{fig:1}} 
\hfil
\subfloat[]{\includegraphics[width=.2\textwidth]{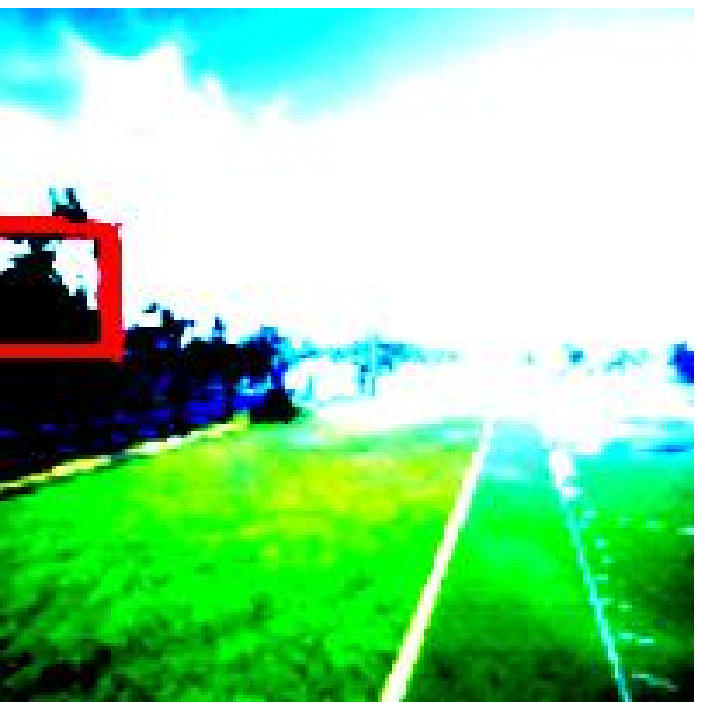} \label{fig:2}}
\hfil
\subfloat[]{\includegraphics[width=.2\textwidth]{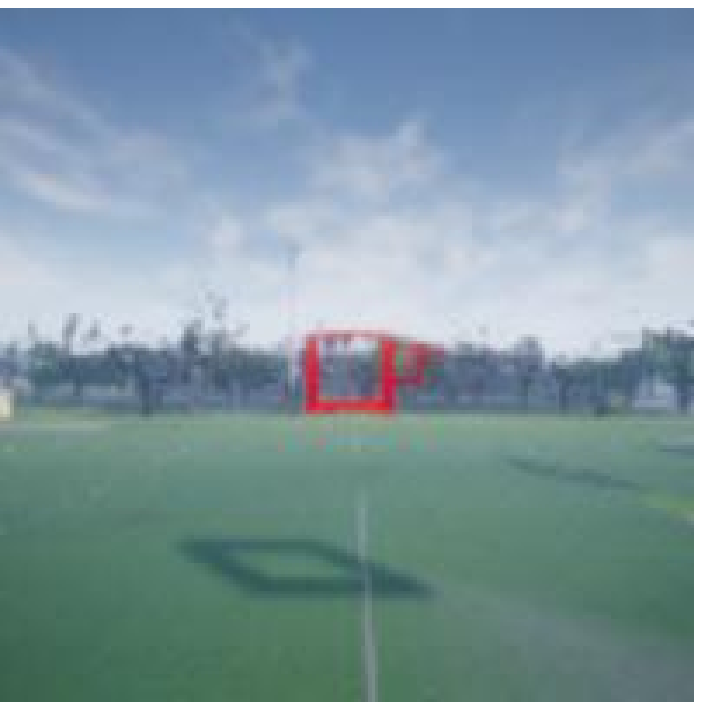} \label{fig:3}}
\hfil
\subfloat[]{\includegraphics[width=.2\textwidth]{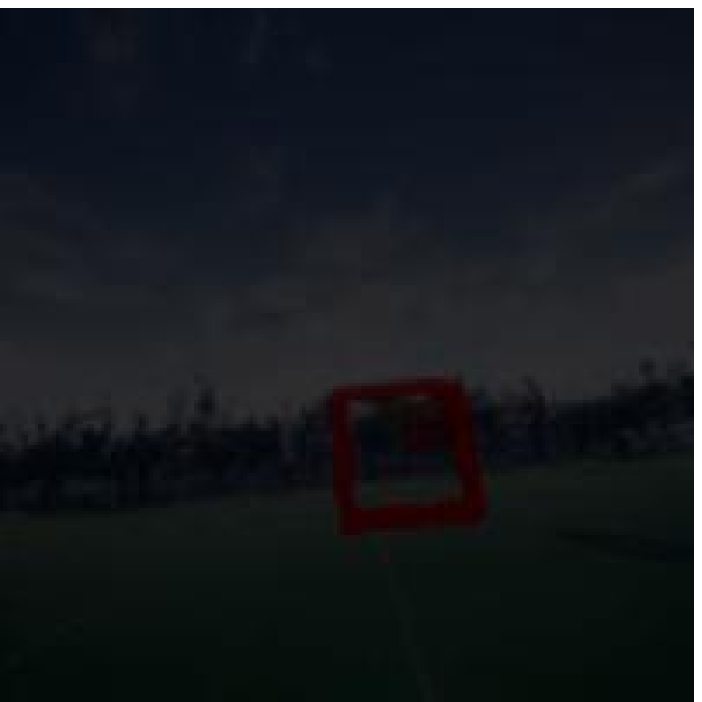} \label{fig:4}}
\caption{Frames in different light conditions in the simulation experiments. See Table~\ref{table:noise} for noise levels for each frame.} 
\label{fig:frames}
\end{figure*}



\begin{figure}
\centering
{\includegraphics[scale=0.12]{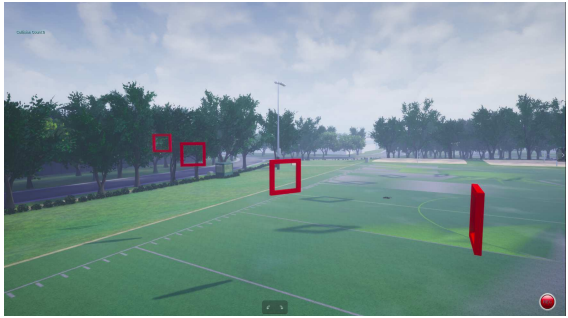}}
\caption{A sample racing track created in AirSim environment.} \label{fig:track}
\end{figure}

We use custom racing tracks with multiple gates in the AirSim environment for the simulation experiments. An example track is shown in Fig.~\ref{fig:track}. We implement the full system in Fig.~\ref{fig:wholesystem}, where the pre-trained DroNet, CovNet (see Section \ref{s:covnet}) and UDS (see Section \ref{s:uds}) are used in harmony to achieve fast completion of the track as fast as possible.

In order to emulate camera noise, drone's virtual camera was exposed to different light/brightness conditions at various timesteps, as displayed in Fig. \ref{fig:frames}, where the noise levels and estimated covariance for each frame is given in Table~\ref{table:noise}. As seen in Table~\ref{table:noise}, frames corrupted with noise are estimated to have higher covariances by CovNet.

We compare the performance of UDS in simulated tracks against using a fixed-strategy for motion planning, independent of CovNet outputs. As can be seen in the Table \ref{table:basari_sure}, six different methods have been compared, four of which are classical motion planning algorithms: minimum speed, minimum acceleration, minimum jerk and minimum jerk full stop. There are two important criteria to evaluate the performance of the declared methods, which are speed and safety. Safety refers to what percentage of the drone's flights have been successfully completed without crashing. Speed indicates how long the drone takes to finish the given track. In order to take the stochasticity of simulations into account, $40$ test flights were conducted under different noise conditions and gate placements. The final results are given as average values with standard deviations in Table \ref{table:basari_sure}. 

In terms of speed, the conventional algorithms, \textit{minimum acceleration}, \textit{minimum jerk} and \textit{minimum jerk full stop}, offers a shorter completion time for a given track. However, from the safety standpoint, all conventional motion planners suffer extensively from perceptual disturbances. Therefore, \textit{minimum acceleration, minimum jerk} and \textit{minimum jerk full stop} methods were only able to complete $60\%, 10\%, 15\%$ of all given flights, respectively. Even the safest traditional motion planner, \textit{minimum speed} has some failed flights. In order to ensure a mixture of safety and speed in agile flights, our approach, \textit{UDS}, is proposed. This approach ensures safety and offers a sufficiently fast flight experience with an average of 1.8 m/s, even if not the fastest one. 

As Table \ref{table:method_dist} shows, when UDS approach is applied, optimal motion planning strategy is predicted considering the circumstances at that moment automatically. While safer and slower methods are selected in noisy conditions, faster planners are preferred in non-noisy situations. Since the classifier model is trained with flight data, this approach does not require any additional code blocks, or heuristics, such as \textit{if-else conditions}. Considering the drone states, the gate pose and the measured co-variance, the drone takes action and safely completes the given track in a relatively short time. With severe perceptual disturbances, the drone enters safe mode and hovers briefly in the air to get a better frame. With this setting, safety is guaranteed at a cost of completion time with a few more seconds.

\begin{table}
\begin{center}
\begin{tabular}{ccc}
\cline{2-3}
\multicolumn{1}{c|}{}           & \multicolumn{1}{c|}{\textbf{Success rate(\%)}} & \multicolumn{1}{c|}{\textbf{Av. Lap time(s)}} \\ \hline
\multicolumn{1}{|l|}{\textbf{UDS}} & \multicolumn{1}{c|}{$100$} & \multicolumn{1}{c|}{$20.15 \pm 2.65$}\\ \hline
\multicolumn{1}{|l|}{\textbf{Minimum  Velocity}} & \multicolumn{1}{c|}{$92.5$} & \multicolumn{1}{c|}{$22.9 \pm 3.7$}\\ \hline
\multicolumn{1}{|l|}{\textbf{Minimum Acceleration}} & \multicolumn{1}{c|}{$60$} & \multicolumn{1}{c|}{$16.2 \pm 2.36$}\\ \hline
\multicolumn{1}{|l|}{\textbf{Minimum Jerk}} & \multicolumn{1}{c|}{$10$} & \multicolumn{1}{c|}{$19.17 \pm 2.12$}\\ \hline
\multicolumn{1}{|l|}{\textbf{Minimum Jerk Full Stop}} & \multicolumn{1}{c|}{$15$} & \multicolumn{1}{c|}{$18.98 \pm 2.56$}\\ \hline
\end{tabular}
\caption{Success rates and average lap completion time for successful flights. Note that lap time is averaged only over successfully completed flights.}
\label{table:basari_sure}
\end{center}
\end{table}
\begin{table}
\begin{center}
  \begin{tabular}{l|c|}
  \cline{2-2}
      & \textbf{Percentage distribution in UDS} \\ \hline
      \multicolumn{1}{|l|}{\textbf{Minimum Velocity}} &  47.8\% \\ \hline
      \multicolumn{1}{|l|}{\textbf{Minimum Acceleration}} & 28.4\% \\ \hline
      \multicolumn{1}{|l|}{\textbf{Minimum Jerk}} &  0.3\% \\ \hline
      \multicolumn{1}{|l|}{\textbf{Minimum Jerk Full Stop}} & 9.5\% \\ \hline
      \multicolumn{1}{|l|}{\textbf{Safe Mode}} & 14.0\% \\
    \hline
  \end{tabular}
  \caption{Distribution of selected planners by UDS.}
  \label{table:method_dist}
  \end{center}
\end{table}

\begin{table}
\begin{center}
  \begin{tabular}{l|c|c|c|c|}
  \cline{2-5}
      & \textbf{Fig \ref{fig:1}} & \textbf{Fig \ref{fig:2}} & \textbf{Fig \ref{fig:3}} & \textbf{Fig \ref{fig:4}} \\ \hline
      \multicolumn{1}{|l|}{\textbf{Brightness}} &  0. & 2.114 & 0. & 3.969\\ \hline
      \multicolumn{1}{|l|}{\textbf{Contrast}} &  0. & 3.811 & 0. & 3.233\\ \hline
      \multicolumn{1}{|l|}{\textbf{Saturation}} & 0. & 0.039 & 0. & 0.0125 \\ \hline
      \multicolumn{1}{|l|}{\textbf{Estimated Covariance}} & 0.114 & 1.15 & 0.141 &0.48\\ \hline
      \multicolumn{1}{|l|}{\textbf{Predicted distance}} & 4.111 & 1.625 & 6.107 &3.66 \\ \hline
  \end{tabular}
  \caption{Noise levels for frames in Fig. \ref{fig:1},\ref{fig:2},\ref{fig:3},\ref{fig:4}.}
  \label{table:noise}
  \end{center}
\end{table}

\subsection{Real World Experiments}

\begin{figure}
    \centering
\includegraphics[scale=0.335]{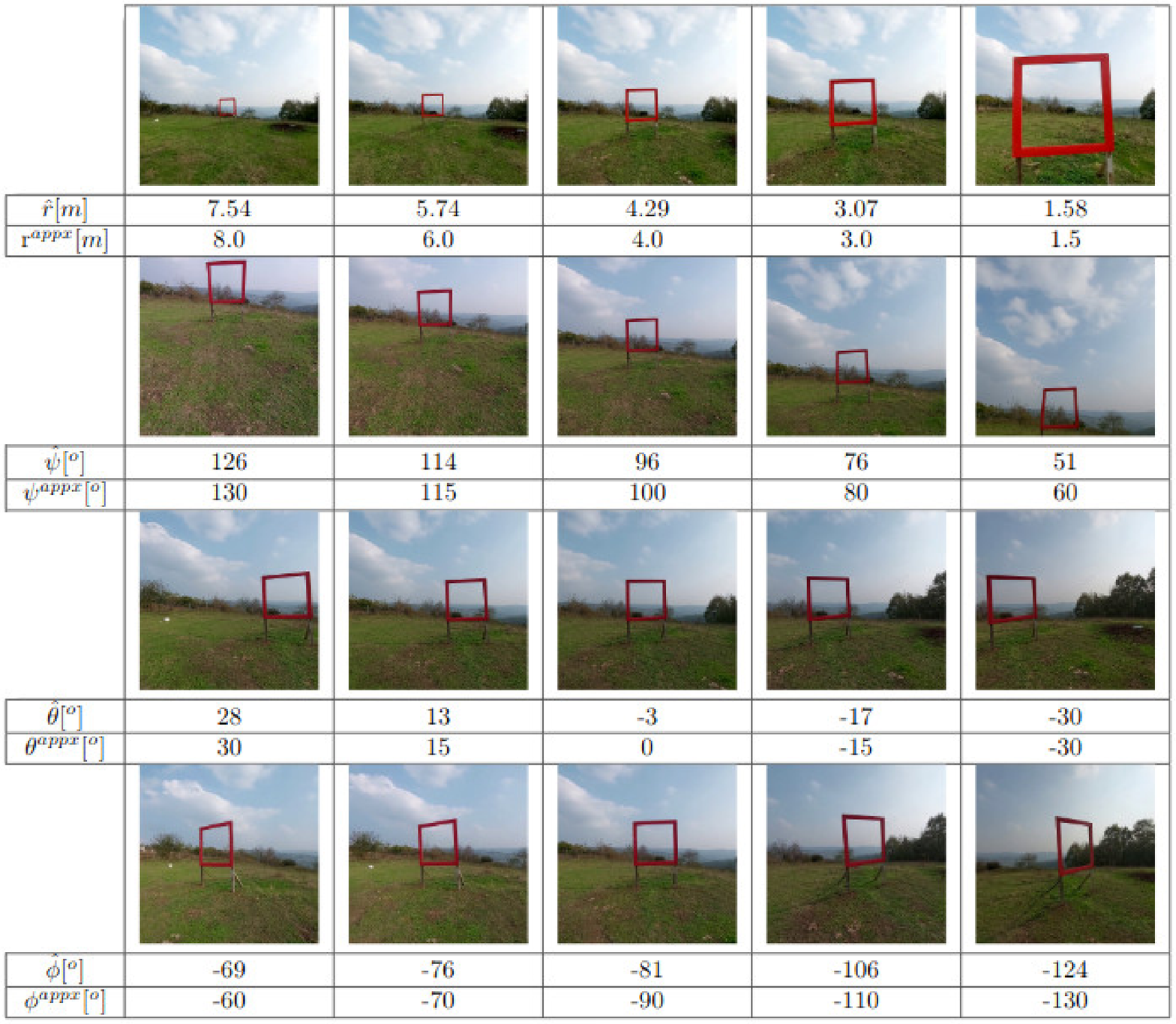}
\caption{DroNet gate pose prediction for real world images taken from different angles and distances.}
    \label{fig:dronetoutput}
\end{figure}

\begin{figure}
    \centering
\includegraphics[scale=0.32]{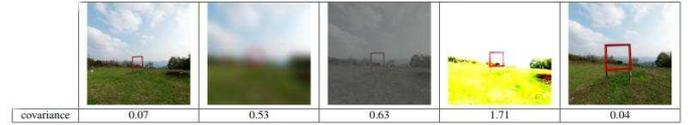}
\caption{CovNet covariance predictions of real world images taken from different conditions/noise levels.}
    \label{fig:covnetoutput}
\end{figure}
For real time experiments, we collect data with a custom drone that consists of F450 frame along with 920KV motor, 8045 carbon fiber propeller, 20A ESC and 4S 5000 mAh Li-Po battery. A pixhawk cube orange with 3 GPS was used as controller board and no additional external sensors were used for the experiments except an RGB camera. Nvidia Jetson Nano was employed as the main computer to run our developed deep learning models. A wi-fi module for communication and a USB camera for perception were connected to Jetson Nano. Image processing and mission planning algorithms  were implemented in Jetson Nano with 15 Hz frequency. Deployed camera operates at 30 Hz frequency. A red square gate with a side length of 1.6 m was constructed, which is similar to the gates used in simulation environment. 
    

Fig.~\ref{fig:dronetoutput}, DroNet gate pose predictions of various images collected by the drone camera is presented.
The first row of shows predicted values Results, whereas the second row shows approximated ground truth values. These results show that DroNet trained in AirSim environment can generalize well to real world images without much fine tuning. Fig.~ \ref{fig:covnetoutput}, shows several real images corrupted by different blur and brightness conditions and their corresponding .~This small-scale test shows that CovNet estimates higher covariance for corrupted images.


\section{Acknowledgements}
This work is supported by the Scientific Research Project Unit (BAP) of Istanbul Technical University, Project Number: MOA-2019-42321 



\FloatBarrier
\bibliographystyle{IEEEtran}
\bibliography{references}

\end{document}